\documentclass[pdflatex,sn-mathphys-num]{sn-jnl}

\usepackage{float}
\usepackage{adjustbox}
\usepackage{graphicx}%
\usepackage{multirow}%
\usepackage{amsmath,amssymb,amsfonts}%
\usepackage{amsthm}%
\usepackage{mathrsfs}%
\usepackage[title]{appendix}%
\usepackage{xcolor}%
\usepackage{textcomp}%
\usepackage{manyfoot}%
\usepackage{booktabs}%
\usepackage[ruled]{algorithm2e}

\usepackage{algpseudocode}%
\usepackage{listings}%
\usepackage{multirow}
\usepackage{subfigure}
\usepackage{booktabs}
\usepackage{hyperref}
\usepackage{amsmath}
\usepackage{array}
\usepackage{amssymb}
\usepackage{mathtools}
\usepackage{amsthm}
\usepackage{extarrows}
\usepackage{diagbox}
\usepackage{bbold}
\usepackage{color}

\definecolor{text}{rgb}{0.4,0.1,0.4}

\theoremstyle{thmstyleone}%
\theoremstyle{thmstyletwo}%

\theoremstyle{thmstylethree}%

\begin{document}

\title[Article Title]{From Generative Engines to Actionable Simulators: The Imperative of Physical Grounding in World Models}


\author*[1]{\fnm{Zhikang} \sur{Chen}}\email{zhikang.chen@eng.ox.ac.uk}

\author[1]{\fnm{Tingting} \sur{Zhu}}

\affil*[1]{\orgdiv{Department of Engineering Science}, \orgname{University of Oxford}, \orgaddress{\city{Oxford}, \country{U.K.}}}



\abstract{
A world model is an AI system that simulates how an environment evolves under actions, enabling planning through imagined futures rather than reactive perception. Current world models, however, suffer from \emph{visual conflation}: the mistaken assumption that high-fidelity video generation implies an understanding of physical and causal dynamics. We show that while modern models excel at predicting pixels, they frequently violate invariant constraints, fail under intervention, and break down in safety-critical decision-making. This survey argues that visual realism is an unreliable proxy for world understanding. Instead, effective world models must encode causal structure, respect domain-specific constraints, and remain stable over long horizons. We propose a reframing of world models as \emph{actionable simulators} rather than visual engines, emphasizing structured 4D interfaces, constraint-aware dynamics, and closed-loop evaluation. Using medical decision-making as an epistemic stress test, where trial-and-error is impossible and errors are irreversible, we demonstrate that a world model’s value is determined not by how realistic its rollouts appear, but by its ability to support counterfactual reasoning, intervention planning, and robust long-horizon foresight.}




\maketitle

\section{Introduction}\label{sec1}

Fundamentally, a world model serves as an artificial intelligence (AI)'s internal representation of the environment, enabling it to simulate future states and the consequences of its actions. World models \citep{ha2018world, matsuo2022deep, zhou2025hermes} have re-emerged as a central abstraction in AI, especially in autonomous driving \citep{wang2024driving, bogdoll2025muvo}, embodied navigation \citep{nie2025wmnav, bar2025navigation}, gaming \citep{ying2025assessing}, and healthcare \citep{mewm}. driven by rapid advances in generative modeling and large-scale representation learning. By learning to predict future observations conditioned on past states and actions, modern models appear increasingly capable of simulating complex environments. However, the current "generative boom" has introduced a profound ambiguity: a growing tendency to equate increasingly realistic video rollouts with genuine world understanding. Here, we must distinguish between two forms of error. Perceptual hallucinations (e.g., blurring textures or inconsistency in lighting) are merely aesthetic failures. However, dynamical hallucinations, where a model generates a plausible-looking video that violates invariant laws, such as a glass shattering before impact or a tumor shrinking without treatment, represent a fundamental failure of world modeling. In safety-critical regimes, the latter is not a glitch; it is a breakdown of causal reasoning.

\begin{figure*}[h!]
    \centering
    \includegraphics[width=0.95\columnwidth]{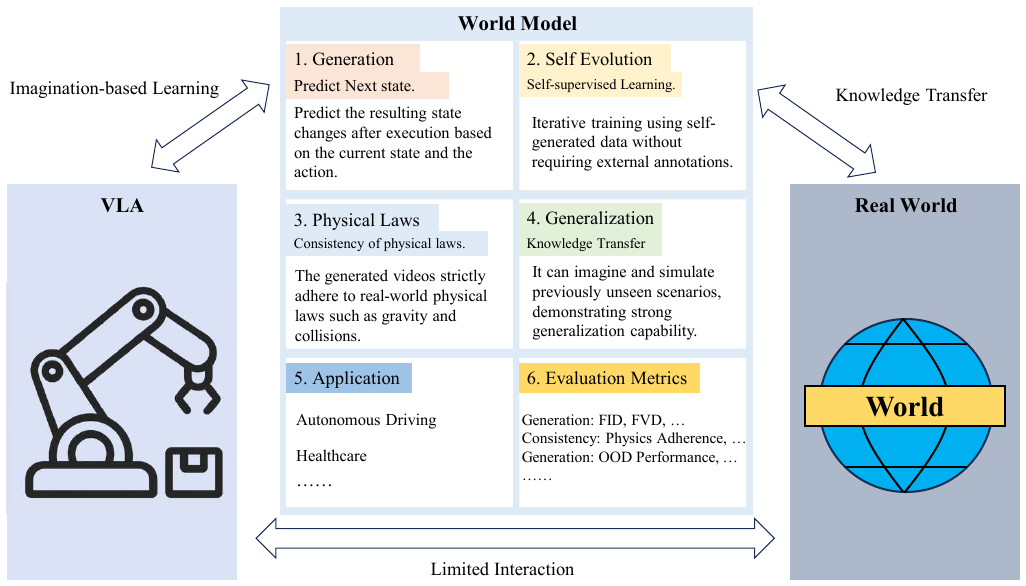}
    \caption{Direct interaction between robots (VLA) and the real world is inherently limited due to safety, cost, and efficiency constraints. A world model enables imagination-based learning by simulating environment dynamics, physical laws, and state transitions in a latent space. Through self-supervised generation, generalization, and evaluation within the world model, the agent can acquire richer knowledge beyond observed data and effectively transfer the learned knowledge and policies to real-world tasks.}
    \label{fig:overall11}
\end{figure*}

In this survey, we argue that visual fidelity is a necessary but insufficient condition for world modeling. Although photorealistic generators excel at rendering the appearance, they often lack the structural depth \citep{zhen20243d}, causal grounding \citep{po2025long, leispartan}, and physical consistency \citep{worldmodelbench, zhan2025phyvllm} required for reliable deployment in long-horizon, safety-critical settings. As illustrated in Figure~\ref{fig:overall11}, the direct interaction between the agents embodied and the real world is inherently limited by the constraints of safety, cost, and data efficiency. World models address this limitation by enabling imagination-based learning: agents can simulate future states, actions, and physical outcomes in a latent space, acquiring knowledge that would be impractical or risky to obtain through real-world experience alone. Crucially, such a simulated experience supports self-evolution, generalization beyond observed data, and knowledge transfer to real-world deployment.

The primary contribution of this work is to synthesize a vast array of recent, disparate advances of the world model field into a unified conceptual framework, highlighting a fundamental shift in the field: the transition from \emph{visual engines} that predict pixels to \emph{actionable simulators} that represent dynamics. Rather than categorizing models by architecture, we organize recent progress around four recurring challenges that define the boundaries of a true world model: 

\textbf{From Generative to Structural Interfaces.} We observe an evolution from 2D pixel-level extrapolation toward structured 4D dynamic meshes, persistent memories, and causal interaction graphs \citep{zhou2025learning, leispartan}. This transition ensures that the world state is not merely rendered, but explicitly exposed for long-horizon reasoning.

\textbf{The Engine of Self-Evolution.} We analyze how world models maintain long-term consistency through closed-loop self-evolution \citep{wang2023robogen, genrl}. By treating generated rollouts as feedback signals to refine internal dynamics, these models move beyond static prediction toward adaptive simulation.

\textbf{The Imperative of Physical Anchoring.} We argue that adaptability without grounding leads to unbounded drift. We evaluate how physics-informed constraints, ranging from explicit differentiable dynamics to implicit intuitive physics, act as a necessary anchor to ensure that self-evolution remains tied to reality \citep{li2025pin, garrido2025intuitive}.

\textbf{Generalization through Structured Imagination.} We discuss the role of world models as data engines in interaction-starved regimes. By learning in uncertainty-aware imagination \citep{whale, diwa}, agents can acquire transferable skills without the cost or risk of real-world trial-and-error.

To demonstrate the criticality of these principles, we examine the medical domain, an environment where counterfactual reasoning is essential and "visual hallucinations" are unacceptable \citep{mewm}. In this high-stakes setting, we show that world modeling is not merely an enhancement but a prerequisite for autonomous decision-making. Finally, we call for a shift in evaluation paradigms, asserting that closed-loop, decision-oriented assessment provides the most principled criterion for distinguishing a predictive world model from a sophisticated generative engine. Through this synthesis, we provide a roadmap for the next generation of world models -- judged not by how they look, but by how reliably they enable us to act.

\section{External Interfaces of World Models: From Generation to Long-Horizon and Causal Futures}

Recent progress in world modeling is inextricably linked to the generative AI boom, where the ability to predict and render future observations acts as the primary \emph{external interface} to a model's latent internal state. Historically, world models have relied on 2D image or video generation for this interface, extrapolating future frames conditioned on past observations and actions. While these models have enabled short-horizon planning in robotics, they encode world state implicitly within pixels, dangerously conflating surface appearance with underlying dynamics. Consequently, purely 2D interfaces face fundamental limitations, specifically loss of object persistence, hallucination in unobserved regions, and temporal drift once objects leave the field of view \citep{zhou2025learning}. These failure modes highlight a critical realization: visual realism alone is not a guarantee that a model has captured the underlying structural laws of the world.


To overcome these limitations, recent work has progressively enriched external interfaces along two complementary axes: \emph{structural explicitness} and \emph{temporal scope}. For Structural Explicitness, modern approaches lift predictions into structured 3D and 4D representations, such as persistent scene memories, point clouds, or dynamic meshes. This shift exposes spatial and geometric regularities that remain stable across viewpoints and time, enabling a significantly more coherent multi-step prediction \citep{zhou2025hermes,zhen2025tesseract}.
SPARTAN \citep{leispartan} introduces a sparse transformer world model that learns time-dependent local causal graphs between entities. By discovering a sparse, state-dependent interaction structure, it allows long video rollouts to remain stable by attending only to causally relevant interactions, mitigating compounding errors.
Complementarily, PoE-World represents world models as products of programmatic experts, composing many small, interpretable causal rules into a coherent long-horizon simulator capable of generalizing from sparse demonstrations \citep{piriyakulkij2025poe}.

Although operating at different levels, these aforementioned approaches share a common insight: reliable long-horizon world modeling requires external interfaces that make temporal abstraction and causal structure explicit, rather than relying solely on implicit frame-to-frame prediction. From this perspective, generative outputs should be viewed not as the definition of a world model, but as structured manifestations. Their value is determined not by visual quality, but by their ability to preserve causality, support intervention, and remain interpretable over long horizons.

\section{Self-Evolution and Long-Term World Consistency}
While structured external interfaces expose the world's underlying geometry and causality, they do not intrinsically guarantee long-term temporal stability \citep{mccloskey1989catastrophic}. Even the most sophisticated world models remain susceptible to fidelity drift, the accumulation of marginal errors that leads to catastrophic divergence from reality once deployed in closed-loop settings. To maintain coherence over extended horizons, models must move beyond static prediction toward active calibration. This necessitates a second axis of progress: \emph{self-evolution}, a paradigm where world models are continually refined through the iterative interplay of generation, imagination, and environmental feedback.

A defining characteristic of self-evolving systems is that generative rollouts serve not just as outputs, but as the primary medium for learning. In these regimes, generated futures act as synthetic training signals that expose modeling inaccuracies. RoboGen \citep{wang2023robogen} exemplifies open-ended simulation by enabling agents to autonomously propose tasks and synthesize new environments. This forces the internal world model to adapt continuously as the distribution of explored scenarios expands. Similarly, imagination-based frameworks like GenRL \citep{genrl} embeds self-evolution directly within the latent dynamics, allowing agents to refine both policies and world-representations entirely through imagined experience, effectively turning imagination into a supervised corrective mechanism. 

Self-evolution also manifests itself through explicit error-driven feedback, where discrepancies between predicted and realized outcomes guide model updates. LLM3 \citep{wang2024llmˆ} demonstrates this in complex task planning. By detecting infeasible plans, such as collisions or kinematic violations, it feeds these data back to recalibrate internal action representations, curbing the cascading errors typical of long-horizon execution. DrEureka \citep{ma2024dreureka} extends this principle to the sim-to-real frontier, iteratively aligning simulated dynamics with real-world performance through automated reward and randomization tuning. Furthermore, CARD \citep{sun2025large} leverages preference feedback for objective-level refinement, stabilizing long-term behavior and preventing the model from "gaming" or exploiting its own predictive inaccuracies. Synthesizing these advances, we observe that self-evolution is less an architectural choice than a fundamental requirement for systemic plasticity. It emerges whenever a world model is embedded in a feedback loop that reuses its own rollouts to challenge and revise its internal assumptions. For world models to transcend short-horizon prediction and succeed in open-ended decision-making, they must remain ``live" entities --  continually updating their understanding in response to the friction between their predictions and the realized world. However, critically, this systemic plasticity bears a sociological risk. If a world model is initialized on biased historical data, the self-evolutionary loop can inadvertently amplify these biases, creating a "feedback loop of reality" where the model simulates and reinforces warped social or clinical trajectories. For example, in healthcare, a model might hallucinate poorer prognoses for underrepresented demographics not based on physiology, but on learned systemic inequities. Thus, physical anchoring must be complemented by normative alignment to ensure simulations remain not just physically possible, but ethically representative.

\section{Self-Evolution with Physical Anchors: Physics-Informed World Models}

The self-evolving mechanisms discussed previously highlight the necessity of closed-loop refinement to preserve long-horizon consistency. However, adaptability in isolation is a double-edged sword. Without robust inductive biases, iterative self-evolution can inadvertently amplify latent modeling errors, leading to ontological drift where a model converges upon states that are internally consistent yet physically impossible. We define this as causal hallucination: a state in which the model’s internal transition function diverges from the laws of physics (e.g., 'forgetting' friction to minimize prediction error). Unlike visual artifacts, causal hallucinations render the model mathematically invalid for control.
In the context of machine intelligence, this drift represents a fundamental misalignment between the learned representation and the underlying causal structure of the environment. In purely data-driven regimes, observational consistency offers no guarantee of adherence to physical laws; minor inaccuracies in latent state estimation can accumulate into catastrophic failures. Consequently, unconstrained self-evolution risks reinforcing spurious correlations. This requires a shift toward \emph{physical anchoring}: using the immutable laws of physics not merely as external guardrails, but as intrinsic inductive biases that shape the manifold of learned representations. 


Recent advances demonstrate that this anchoring can be achieved through a spectrum of explicit and implicit mechanisms. At the explicit end of the spectrum, models such as PIN-WM \citep{li2025pin} embeds differentiable rigid-body dynamics directly into the computational graph. By integrating parameter identification into the learning objective, this approach constrains the optimization landscape, ensuring that the model searches only within the subspace of physically interpretable quantities. This prevents the ``shortcut learning" typical of pure generative models, forcing the network to solve for dynamics rather than texture statistics. Physical grounding can also be enforced implicitly through the cross-consistency of sensory modalities. RoboScape \citep{shang2025roboscape} demonstrates that jointly optimizing video, depth, and keypoint dynamics acts as a regularizer. Here, the physical structure of the world is not hard-coded but inferred from the geometric agreement between different views of the same underlying state, stabilizing long-horizon predictions.


Perhaps most promising for general-purpose intelligence is the emergence of intuitive physics within self-supervised frameworks. WISA \citep{wang2025wisa} injects physical priors into video diffusion, constraining the temporal evolution of futures generated. V-JEPA \citep{garrido2025intuitive} advances this by operating entirely in latent space. It illustrates that principles such as object permanence and spatiotemporal continuity can emerge from joint-embedding predictive objectives. By penalizing physically nonsensical transitions in abstract representation space, it achieves grounding without the need for symbolic physics engines.



Synthesizing these perspectives, we argue that the next generation of world models must resolve the dialectic between self-evolution (plasticity) and physical anchoring (structure). Without evolution, representations remain static and brittle; without physical structure, evolution lacks a reality check. Integration of these paradigms marks a step toward embodied intelligence, where models refine their understanding of complex tasks while remaining tied to the immutable laws of the physical world.

\section{Generalization under Limited Interaction: Learning in Imagination}

A primary requirement of world models is their capacity to catalyze generalization \citep{liu2021towards, chen2024neural, kachaev2025don} beyond the confines of observed interactions, particularly in ``interaction-starved" regimes where real-world exploration is prohibitively expensive, risky, or ethically constrained. In such settings, endemic to robotics and safety-critical systems, the locus of learning shifts from trial-and-error in the physical environment to \emph{imagination}: the strategic use of a learned simulator to plan, adapt, and evolve. Here, the burden of generalization is transferred from the quantity of raw data to the structural integrity and constraints of the world model itself. However, naïvely learning from hallucinated trajectories is dangerous, if imagined rollouts are unconstrained by reality, they risk reinforcing spurious correlations and allowing agents to exploit model artifacts rather than acquire transferable skills.


Recent scholarship highlights that reliable learning in imagination requires a delicate balance between expressivity and principled regularization. GenRL \citep{genrl} demonstrates that by aligning latent dynamics with multimodal vision-language representations, agents can synthesize and learn novel tasks entirely within imagination, achieving data-free adaptation. DiWA \citep{diwa}: leverages diffusion-based world models as latent MDPs for offline policy fine-tuning, significantly reducing the requisite real-world interaction. To prevent agents from ``gaming" their own imagination, uncertainty-aware mechanisms have become indispensable. WHALE \citep{whale} utilizes behavior-conditioning to bound policy-induced distribution shifts, ensuring the agent stays within the model's reliable zone. WM-VAE \citep{wmvae} explicitly penalizes out-of-distribution states during the planning phase, acting as a safeguard against novelty-seeking in dangerous or undefined regions.

We further observe that generalization is profoundly enhanced when imagination operates over structured, semantically dense representations rather than raw pixels. By abstracting away embodiment-specific details through object-centric states or 3D geometry, as seen in 3DFlowAction \citep{flow} and DyWA \citep{dywa}, world models facilitate seamless cross-task and cross-platform transfer. 
Beyond offline training, techniques such as \citep{lps, reoi} allow for the preemptive detection of implausible futures at inference time, enabling real-time re-planning under distributional drift. At scale, systems such as EmbodieDreamer \citep{embodiedreamer} and GigaBrain-0 \citep{gigabrain} herald a paradigm where world models function as autonomous data engines, generating vast, physically grounded synthetic experiences, effectively decoupling agent intelligence from the scarcity of real-world data.

Synthesizing these trends, we argue that generalization in interaction-limited regimes is not a byproduct of larger policy architectures, but an emergent property of world models acting as constrained internal learning environments. For imagination to be a viable surrogate for reality, it must be structured, verified, and acutely aware of its own epistemic limits.

\section{World Models in Medical Settings: The Ultimate Epistemic Stress Test}

Medical scenarios \citep{alazab2022digital, kaul2023role, meijer2023digital, chaparro2025technological, ringeval2025advancing, de2025could} represent perhaps the most formidable frontier for intelligent systems. Unlike standard robotics benchmarks, clinical decision-making is defined by sparse and biased observations, irreversible interventions, and severe, often fatal, consequences for failure. In this domain, the demands placed on imagination-based learning transcend optimization metrics and enter the realm of \emph{epistemic responsibility}. Medical settings thus constitute the ultimate stress test for world models: they reveal whether learned dynamics can support rigorous counterfactual reasoning and long-horizon foresight when trial-and-error is ethically and practically impossible.

Crucially, the constraints governing medical environments differ fundamentally from those in classical physical control systems. Although robotic manipulation benefits from explicit rigid-body dynamics and kinematic laws, internal medicine is shaped instead by pathophysiological processes, homeostatic regulation, and biochemical interactions. As such, the appropriate objective is not a narrowly construed ``physics-informed'' model, but rather a domain-constrained, biologically grounded world model. Here, the governing ``laws'' are clinical in nature: physiological viability ranges (e.g., blood pressure or oxygen saturation bounds), pharmacokinetic and pharmacodynamic rules, disease progression priors, and known causal asymmetries between intervention and outcome. These constraints serve an analogous role to physical laws in robotics, delimiting the space of plausible futures and ruling out biologically impossible trajectories.

A paradigmatic example is the Medical World Model (MeWM) \citep{mewm}, which reformulates clinical decision-making as an action-conditioned stochastic dynamical system. By decomposing the problem into a treatment policy, a disease dynamics simulator, and a survival-based evaluation module, MeWM enables the synthesis of counterfactual post-treatment trajectories—such as tumor progression or regression—that cannot be directly observed or experimentally tested. This illustrates a defining imperative of medical world models: because disease evolution is stochastic, partially observed, and long-horizon, forward simulation becomes the primary vehicle for informed reasoning. Importantly, such imagination must remain tightly anchored to biological and pathological constraints. In medicine, the definition of hallucination shifts from the kinematic to the physiological. A clinical hallucination occurs when a generated trajectory violates homeostatic bounds—for example, a tumor shrinking spontaneously without treatment. Thus, unconstrained generative rollouts are not merely inaccurate but clinically dangerous.

Beyond internal disease modeling, world model principles are equally critical for embodied medical systems operating under extreme safety requirements. RoboNurse-VLA \citep{robonurse}, for instance, demonstrates how vision-language-action models can assist in surgical instrument handover despite limited data, high precision demands, and near-zero tolerance for error. By leveraging structured priors about human intent, task semantics, and environmental affordances, the system generalizes to unseen tools and gestures—embodying the core promise of world models: maintaining reliability under distributional shift through structured internal simulation.

Synthesizing these developments, medicine should not be viewed as an edge case for world models, but rather as the domain that most sharply exposes their foundational assumptions. Without biologically grounded world models, medical AI remains confined to reactive, correlational inference. With them, it can mature into a genuinely decision-oriented partner. This clarifies the fundamental value of world modeling in healthcare: providing a stable, constrained imagination for environments where the cost of a mistaken rollout is a human life.

\section{Evaluation: When Does a Model Qualify as a World Model?}

Despite the rapid advances in generative modeling, the evaluation of world models remains a central and unresolved challenge. This progress raises an unavoidable ontological question: what qualifies a model as a world model, rather than a sophisticated generator? If visual realism and open-loop prediction accuracy are insufficient proxies for decision-making reliability, evaluation must be reframed around interaction, intervention, and long-horizon consequences.

Traditional benchmarks often inherit protocols from image and video synthesis, emphasizing perceptual fidelity, temporal smoothness, and distributional similarity. However, these criteria are increasingly recognized as inadequate for assessing whether a model has internalized the governing structure of an environment. A visually compelling rollout may still violate invariant constraints, ignore causal actions, or collapse under closed-loop control. Empirically, high scores on perceptual metrics such as FID or FVD correlate weakly with downstream planning or control performance, motivating a shift toward decision-centric and structure-aware evaluation.

One prominent response to this gap is the introduction of benchmarks explicitly targeting world-model-specific properties. WorldModelBench \citep{worldmodelbench}, for example, evaluates instruction following, commonsense reasoning, and physical adherence, revealing systematic failures, such as violations of gravity or mass conservation, that remain invisible to standard visual metrics. While such benchmarks often rely on human-annotated judgments of plausibility, their primary contribution is diagnostic rather than scalable: they expose failure modes and establish qualitative upper bounds on model understanding, rather than serving as the sole mechanism for routine evaluation or training-time feedback.

To address scalability and reproducibility concerns, recent work increasingly explores automated and constraint-based metrics that approximate physical or causal adherence without human supervision. One class of methods measures violations of known invariants, such as energy conservation, object permanence, contact consistency, or monotonicity constraints, directly from generated rollouts. Another leverages inverse dynamics or action inference: if a model’s predicted future states do not admit a plausible action sequence under known control limits, the rollout is deemed inconsistent. These signals can be computed automatically, are amenable to large-scale benchmarking, and in some cases provide differentiable loss terms that can be incorporated during training.

Nevertheless, it is important to acknowledge a fundamental limitation: unlike visual realism, there is currently no universally accepted, task-agnostic analogue of FVD for physical or causal correctness. “Physics adherence” and “commonsense plausibility” remain composite notions that resist full reduction to a single scalar metric. We view this not as a weakness of individual benchmarks, but as an open challenge for the field. Establishing standardized, automated measures of invariant violation and causal consistency is a prerequisite for scalable progress in world modeling.

A complementary and increasingly influential evaluation paradigm bypasses explicit physics metrics altogether by assessing world models through their utility in closed-loop decision-making. Rather than scoring predictions in isolation, this approach treats the learned world as an interactive surrogate environment. WorldEval \citep{worldeval} demonstrates that executing robotic policies within a learned world model yields performance rankings that strongly correlate with real-world deployment outcomes, offering a scalable alternative to physical testing. Similarly, WorldGym \citep{worldgym} formalizes world models as policy evaluation environments, using Monte Carlo rollouts and task-conditioned rewards to assess generalization across diverse scenarios. World-in-World \citep{worldinworld} further shows that in embodied settings, controllability and action fidelity are far more predictive of success than pixel-level accuracy.

Taken together, these developments point to a unifying principle: world models should be evaluated not by how realistic they appear, but by how reliably they support counterfactual reasoning, constraint satisfaction, and long-horizon interaction. From this perspective, evaluation becomes an operational definition. Models that excel on perceptual metrics yet fail under automated constraint checks or closed-loop control remain advanced generators; only those that sustain actionable, causally consistent rollouts merit the designation of true world models.

Based on a synthesis of recent work, we summarize commonly adopted evaluation metrics along five complementary dimensions. First, perceptual generation quality is measured using distributional and similarity metrics such as FID/FVD, LPIPS/SSIM, CLIPScore, and temporal coherence benchmarks (e.g., VBench), capturing visual realism and short-range consistency (Table~\ref{1}). Second, physical and commonsense consistency is assessed via invariant-violation rates, physics adherence scores, and automated plausibility checks that penalize impossible dynamics (Table~\ref{2}). Third, language and multimodal grounding evaluates whether instructions, entities, and relations are correctly aligned with world states, using instruction-following accuracy, question answering, and counterfactual reasoning benchmarks (Table~\ref{3}). Fourth, task- and decision-oriented metrics assess utility for planning and control, including task success rate, policy return, sim-to-real correlation, and safety-related failure rates (Table~\ref{4}). Finally, in domain-specific settings such as healthcare, evaluation incorporates clinically grounded and expert-informed criteria, including anatomical consistency, treatment-conditioned disease progression accuracy, risk prediction error, and survival ranking metrics (Tables~\ref{5} and~\ref{6}). Collectively, these dimensions reflect a decisive shift from purely generative evaluation toward assessing world models as predictive, causal, and decision-support systems.

\begin{table}[t!]
\centering
\caption{Image and Video Generation Metrics}
\label{1}
\begin{tabular}{l|l|l}
\hline
\textbf{Metric} & \textbf{Evaluation Content} & \textbf{Description} \\
\hline
FID / FVD & Distribution similarity to real data & Measures visual realism only. \\
LPIPS / SSIM & Perceptual similarity & Captures frame-level fidelity. \\
CLIPScore & Text--visual alignment & Evaluates semantic consistency. \\
VBench & Temporal consistency & Assesses video coherence. \\
\hline
\end{tabular}
\end{table}

\begin{table}[t!]
\centering
\caption{Physical and Commonsense Consistency Metrics}
\label{2}
\begin{tabular}{l|l|l}
\hline
\textbf{Metric} & \textbf{Evaluation Content} & \textbf{Description} \\
\hline
Physics Adherence & Physical law compliance & Detects physics violations. \\
Commonsense Score & Logical plausibility & Evaluates everyday reasoning. \\
Violation Rate & Invalid events frequency & Penalizes impossible outcomes. \\
\hline
\end{tabular}
\end{table}

\begin{table}[t!]
\centering
\caption{Language and Multimodal Alignment Metrics}
\label{3}
\begin{tabular}{l|l|l}
\hline
\textbf{Metric} & \textbf{Evaluation Content} & \textbf{Description} \\
\hline
Instruction Following & Action compliance & Checks instruction execution. \\
QA / What-if Accuracy & Counterfactual reasoning & Tests causal understanding. \\
Grounding Accuracy & Language--entity alignment & Ensures spatial grounding. \\
\hline
\end{tabular}
\end{table}

\begin{table}[t!]
\centering
\caption{Task and Decision-Oriented Metrics}
\label{4}
\begin{tabular}{l|l|l}
\hline
\textbf{Metric} & \textbf{Evaluation Content} & \textbf{Description} \\
\hline
Task Success Rate & Goal completion & Primary decision metric. \\
Policy Return & Long-horizon reward & Measures planning quality. \\
Real-world Correlation & Sim-to-real consistency & Validates proxy fidelity. \\
Safety Failure Rate & Unsafe behaviors & Measures risk. \\
\hline
\end{tabular}
\end{table}

\begin{table}[t!]
\centering
\caption{Medical Image Generation Metrics}
\label{5}
\begin{tabular}{l|l|l}
\hline
\textbf{Metric} & \textbf{Evaluation Content} & \textbf{Description} \\
\hline
Radiologist Turing Test & Expert realism judgment & Assesses clinical realism. \\
Medical FID / LPIPS & Image similarity & Quantifies visual fidelity. \\
Anatomical Consistency & Structural correctness & Preserves anatomy. \\
\hline
\end{tabular}
\end{table}

\begin{table}[t!]
\centering
\caption{Medical Decision and Dynamics Metrics}
\label{6}
\begin{tabular}{l|l|l}
\hline
\textbf{Metric} & \textbf{Evaluation Content} & \textbf{Description} \\
\hline
Action-conditioned Progression & Treatment response & Tests causal disease modeling. \\
F1 / Precision / Recall & Protocol accuracy & Measures decision quality. \\
Jaccard Index & Protocol overlap & Penalizes redundancy. \\
Risk Score MSE & Survival prediction error & Evaluates prognosis accuracy. \\
C-index & Risk ranking & Measures discrimination. \\
\hline
\end{tabular}
\end{table}

\section{Conclusion: The Imagination Engine of General Intelligence}

The rapid ascent of generative AI has brought world modeling to a critical juncture. As this survey has demonstrated, the field is undergoing a fundamental shift: moving away from visual engines that prioritize pixel-level fidelity toward actionable simulators that represent the underlying physical and causal laws of reality. By synthesizing recent breakthroughs in structured 4D interfaces, self-evolving loops, and physical anchoring, we have clarified a pivotal distinction: a true world model is defined not by how realistically it predicts the future, but by how reliably it supports intervention, reasoning, and decision-making.

Our analysis reveals that the generative boom must be grounded in both structural and physical realities. Whether through explicit geometry, sparse causal graphs, or programmatic rules, the evolution of world models is increasingly characterized by a move toward explicitness. This evolution is perhaps most urgent in safety-critical domains such as medicine, where the cost of visual hallucinations is a human life. In such high-stakes environments, world models serve as the essential crucible for counterfactual reasoning, transforming AI from a reactive observer into a predictive partner.

Looking ahead, the ultimate litmus test for world models will lie in their closed-loop utility. As evaluation paradigms shift from open-loop metrics toward task-centric assessments like WorldEval, the community will begin to distinguish between models that merely look right and those that are physically right. Ultimately, world models represent the ``imagination engine" of AGI. By providing a stable, grounded, and self-correcting internal environment for agents to learn and plan, they decouple intelligence from the scarcity of real-world data. As we move toward this future, the principles of causality, physics, and self-evolution will remain the north star, guiding the development of models that do not merely hallucinate possible futures, but truly understand the laws that govern them.

\bibliography{sn-bibliography}

\end{document}